\documentclass{article}
\usepackage{spconf,amsmath,graphicx}
\usepackage{booktabs}
\usepackage{amsmath}
\usepackage{amsthm}
\usepackage{booktabs}
\usepackage{algorithm}
\usepackage{algorithmic}
\usepackage[switch]{lineno}
\usepackage{makecell}
\usepackage{adjustbox}
\usepackage{mathtools}
\usepackage{amsmath}
\usepackage{lipsum}
\usepackage{wrapfig}
\usepackage{graphicx}
\usepackage{tablefootnote}
\usepackage{amsfonts}
\usepackage{multirow}
\usepackage{hyperref}
\usepackage{url}            

\title{TranSentence: Speech-to-speech Translation via Language-agnostic Sentence-level Speech Encoding without Language-parallel Data}
\name{Seung-Bin Kim, Sang-Hoon Lee, Seong-Whan Lee$^{\dagger}$\thanks{$^\dagger$Corresponding author}}
\address{Department of Artificial Intelligence, Korea University, Seoul, Korea}

\begin{document}
\ninept
\maketitle
\begin{abstract}
Although there has been significant advancement in the field of speech-to-speech translation, conventional models still require language-parallel speech data between the source and target languages for training. In this paper, we introduce TranSentence, a novel speech-to-speech translation without language-parallel speech data. To achieve this, we first adopt a language-agnostic sentence-level speech encoding that captures the semantic information of speech, irrespective of language. We then train our model to generate speech based on the encoded embedding obtained from a language-agnostic sentence-level speech encoder that is pre-trained with various languages. With this method, despite training exclusively on the target language's monolingual data, we can generate target language speech in the inference stage using language-agnostic speech embedding from the source language speech. Furthermore, we extend TranSentence to multilingual speech-to-speech translation. The experimental results demonstrate that TranSentence is superior to other models.
\end{abstract}


\section{Introduction}
\label{sec:intro}
Speech-to-speech translation (S2ST) aims to convert speech from one language into speech in another language. 
The conventional approaches, cascaded methods \cite{599557, 1597243, 10096183}, obtain a translated text through automatic speech recognition with machine translation, or speech-to-text translation. The translated text is then synthesized into speech using text-to-speech synthesis. 
However, these methods have limitations, such as the loss of characteristics like the prosody of the original speech, the problem of error propagation, and the complex pipeline where each model must be trained independently. 

Recently, there has been significant advancement in the field of direct speech-to-speech translation, with research predominantly branching into two main directions. 
The first approach directly predicts the Mel-spectrogram. Translatotron \cite{jia19_interspeech} employs an attention-based sequence-to-sequence model, training its hidden representation to contain phoneme information through auxiliary tasks. Translatotron 2 \cite{jia2022translatotron} connects a speech encoder, linguistic decoder, and acoustic synthesizer with a single attention module. 

The second approach leverages discrete units \cite{9585401} clustered from representations of a self-supervised trained speech encoder as the output for the decoder. For the first time, \cite{lee-etal-2022-direct} incorporated discrete speech units into S2ST, and the predicted units were transformed into audio waveforms through a unit vocoder \cite{polyak21_interspeech}. However, the previous model relied on text to achieve its performance, a speech normalization \cite{lee-etal-2022-textless}, data augmentation \cite{10095578}, and speech quantizer \cite{10096797} are proposed to facilitate S2ST without text. Meanwhile, \cite{inaguma-etal-2023-unity, 10095616} improve speech-to-unit translation by using additional text with two-pass decoding \cite{inaguma-etal-2023-unity}. 

Although there have been notable advancements in the S2ST system, training such a system still necessitates language-parallel data. To address this, recent approaches have been proposed to train the S2ST System utilizing non-parallel data. \cite{diwan2023unit} introduces a method that carries out unsupervised back-translation by bootstrapping off utterance pairs. These translated utterance pairs are generated at the word level by aligning within a monolingual text embedding space. Translatotron 3 \cite{nachmani2023translatotron} also proposed a method that employs unsupervised embedding word mapping for back-translation. 

In this paper, we introduce TranSentence, an S2ST system without language-parallel speech data. As a foundational step, we adopt a language-agnostic speech encoder designed to capture only the sentence-level information from speech, irrespective of its language. This encoder is pre-trained with more accessible datasets from machine translation, speech-to-text translation, and speech-text. During training, we exclusively utilize data only from the target language. Initially, using the pre-trained language-agnostic speech encoder, we encode the target language speech into language-agnostic sentence-level speech embedding. Subsequently, the model is trained to reconstruct the original speech in the target language based on this speech embedding. Given that the speech embedding resides in a consistent embedding space irrespective of language, in the inference stage, we encode the speech from the source language into speech embedding. Subsequently, the target speech is generated based on this speech embedding. 

\begin{figure*}[!t]
  \centering
\centerline{\includegraphics[width=0.99\textwidth]{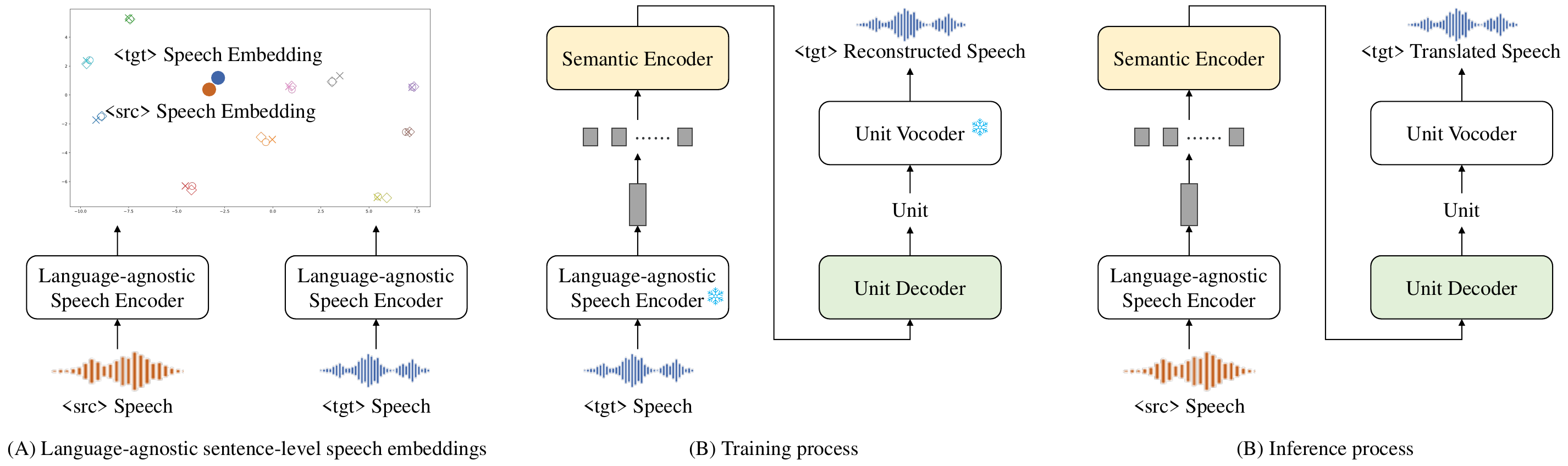}}
\caption{Overview of the TranSentence. The pre-trained language-agnostic sentence-level speech encoder generates speech embedding, and based on this, TranSentence is trained to reconstruct speech in the target (tgt) language. During the inference process, TranSentence generates speech in the target language from the speech embedding of the source (src) language speech.}
\label{f_overview} 
\end{figure*}

Furthermore, we introduce feature expansion, enhancing the ability to generate speech from sentence representation. The experimental results demonstrate that our model outperforms the comparison model. Additionally, we seamlessly expand our model to multilingual speech-to-speech translation. 
Our contributions are summarized as follows:
\begin{itemize}
    \item We introduce TranSentence, a method to train a speech-to-speech translation system without language-parallel speech data.
    \item To the best of our knowledge, this is the first study to utilize language-agnostic sentence-level speech encoding for speech-to-speech translation. 
    \item We also propose modeling methods to generate speech from speech embedding through feature expansion.
    \item The experimental results exhibit that our method  can execute speech-to-speech translation without parallel language speech data, surpassing the other comparative model. 
    \item TranSentence can be expanded to multilingual speech-to-speech translation. Audio samples and code are available at \url{https://sb-kim-prml.github.io/TranSentence}.
\end{itemize}

\section{Related Work}
\label{sec:related}
\subsection{Speech-to-unit translation}
Our method is based on the Transformer-based sequence-to-sequence speech-to-unit translation (S2UT) model \cite{lee-etal-2022-direct}. In the S2UT system, for efficient content learning, the decoder generates speech discrete units instead of mel-spectrograms or audio waveforms. These generated discrete units are then transformed into audio waveforms through a unit vocoder. 
For extracting target speech discrete units, we adopt HuBERT \cite{9585401}, which is the self-supervised approach for speech representation learning and clustering. We adopt unit-to-waveform model \cite{lee-etal-2022-direct, polyak21_interspeech} based on HiFi-GAN \cite{NEURIPS2020_c5d73680}. 
Additionally, we apply a unit reduction strategy for efficient decoding and better translation \cite{lee-etal-2022-direct}. Because this strategy requires the prediction and extension of each unit's duration, a duration prediction module is incorporated.
Since these approaches predict units of pronunciation information rather than higher-dimensional acoustic features, it facilitates easier training. Additionally, a benefit of this method is that it can extract discrete units from audio without text labels and use them as training data. In this regard, We adopt the S2UT system as a baseline to enable learning without text.
\vspace{0.1cm}

\subsection{Language-agnostic sentence embeddings}
Sentence embeddings have been actively researched for various purposes in the text domain. Based on these researches, sentence embeddings are introduced in speech mining to generate language-pair data. SpeechMatrix \cite{duquenne-etal-2023-speechmatrix} extended the capabilities of the pre-trained language-agnostic sentence representations encoding (LASER) from the text domain to the speech domain. They train a speech encoder leveraging LASER, which is pre-trained in various languages, to capture the semantic information of speech irrespective of language. During training, they utilize speech-text pair data and speech-text translation pair data, which are more obtainable than language-pair speech data. This approach not only guides the speech encoder to map both speech and text into a unified semantic embedding space but also enables us to encode the meaning of speech without relying on language-parallel speech data. We adopt this pre-trained speech encoder as a language-agnostic sentence-level speech encoder in our model, allowing us to train an S2ST system without language-parallel speech data.

\section{Method}
In this paper, we propose a speech-to-speech translation system without parallel speech data between the source and target languages, as illustrated in Fig. \ref{f_overview}. To address this, we adopt a pre-trained language-agnostic speech encoder capable of capturing the semantic information of speech irrespective of language. Our model is a Transformer-based sequence-to-sequence model, with a decoder designed to predict discrete units from the sentence representation. To concentrate on validating the capability of generating speech from language-agnostic sentence-level speech embedding, we do not use additional tasks. Under the assumption that the representation encoded by the language-agnostic speech encoder is independent of language, we train our model using only target language data without language-parallel data. In the inference phase, we can predict the units of the target language from the speech of the source language because the representation encoded from the speech of the source language also contains only the semantic information of the speech, irrespective of language. Further details are presented in the following subsections. 
\vspace{1mm}

\subsection{Language-agnostic sentence-level speech encoder}
We adopt the pre-trained speech encoder from SpeechMatrix \cite{duquenne-etal-2023-speechmatrix} as the language-agnostic sentence-level speech encoder for our model. The speech encoder is based on the XLS-R \cite{babu22_interspeech, lee2022hierspeech} architecture. The input speech is encoded through the speech encoder, and the encoder output is max-pooled into a fixed-size representation to capture the meaning of the speech. Consequently, this single-vector speech embedding contains the semantic information of the speech irrespective of language.
\vspace{1mm}

\subsection{Feature expansion}
Our task involves predicting longer speech units from a single-vector speech embedding, which presents challenges in calculating attention in the decoder. Therefore, we aim to facilitate the training of the attention module by enhancing the alignment between the speech embedding and speech units through the feature expansion of the speech embedding. 
For the feature expansion, we introduce a semantic encoder. Firstly, we divide the speech embedding into several sub-embeddings. As a result, the single-vector speech embedding is transformed into a representation composed of multiple frames. Subsequently, the speech sub-embeddings are refined through a semantic encoder to better capture semantic information. 
\vspace{0.1cm}

\subsection{Discrete unit decoder}
The discrete unit decoder is a stack of Transformer blocks and predicts units based on hidden representation from semantic encoder as context information with an attention mechanism. 
Given that the target unit sequence is discrete, we employ cross-entropy loss with label smoothing as our objective function as follows:
\begin{equation}
    \mathcal{L}_{CE}(y,\hat{y}) = -\sum_{i=1}^{N}{y_i}\log{\hat{y}_i},
\end{equation}
\begin{equation}
    \mathcal{L}(y,\hat{y}) = (1-\alpha)\mathcal{L}_{CE}(y,\hat{y}) + \alpha\mathcal{L}_{CE}(u,\hat{y}),
\end{equation}
where $N$, $\alpha$, and $u$ denote the number of tokens including units, label smoothing parameter, and uniform distribution, respectively.
During the inference phase, units are predicted autoregressively. For multilingual speech-to-speech translation with multiple target languages, an additional language tokens are used, unlike when there's just one target language. 
During the inference phase of multilingual speech-to-speech translation, by beginning of sequence with the token corresponding to the desired target language, it is possible to generate the speech discrete units specific to that language. 
\vspace{0.1cm}

\section{Experiments}
\subsection{Datasets}
We employ the training set from Common Voice 11 \cite{ardila-etal-2020-common} as our training data. Among them, we utilized the speech datasets of three languages: English (En), Spanish (Es), and French (Fr). Since Spanish dataset has fewer utterances, totaling 230k, compared to the other languages, we balanced the data volume by randomly sampling 230k utterances each from the English and French train set. 
For the evaluation, we employed the Spanish-English and French-English pair datasets from CVSS-C dataset \cite{jia-etal-2022-cvss}, which are derived from the CoVoST 2 speech-to-text translation corpus. In this dataset, the target text is converted into speech using a text-to-speech system. All audio files were downsampled to a sampling rate of 16,000 Hz.
\vspace{0.1cm}

\subsection{Training setup}
We train TranSentence for 400k steps using the Adam optimizer with $\beta_1=0.9$, $\beta_2=0.98$, and $\epsilon=10^{-8}$ with two NVIDIA RTX A6000 GPUs. For the learning rate scheduling, we use inverse square root scheduler with 10k warm-up steps. We set the initial learning rate at $10^{-7}$ and adjusted it to $0.0005$. The label smoothing parameter $\alpha$ is set to 0.2.
\vspace{0.1cm}

\subsection{Implementation details}
The language-agnostic sentence-level speech encoder consists of a convolutional feature encoder and 48 layers of Transformer blocks with 1,920 embedding dimensions, 7,680 feed-forward network embedding dimensions, and 16 attention heads. We use provided weights for speech encoder from \cite{duquenne-etal-2023-speechmatrix}. The semantic encoder consists of two layer of convolution layer. The discrete unit decoder consist of six layers of Transformer blocks with 256 embedding dimensions, 2,048 feed-forward network embedding dimensions, and four attention heads.
For the unit extraction and vocoder, we use the provided weights of mHuBERT and unit-based HiFi-GAN vocoder from \cite{lee-etal-2022-textless}. 
\vspace{0.1cm}

\subsection{Evaluation metric}
To evaluate the quality of the translation, we measure the BLEU scores. First, we use open-source automatic speech recognition \cite{NEURIPS2020_92d1e1eb, conneau21_interspeech} to transcribe the translated speech into text, and then performed text normalization. After that, we calculated the BLEU scores with the reference text using SacreBLEU. 
\vspace{0.1cm}

\begin{table}[!t]
  \caption{BLEU comparison on CVSS-C dataset. 
  $^\clubsuit$Results from \cite{nachmani2023translatotron}, $^\Diamond$results from \cite{jia-etal-2022-cvss}, $^\spadesuit$results from \cite{huang2023transpeech}, and $^\triangle$results from \cite{inaguma-etal-2023-unity}. S2UT$^\dagger$ is the trained model with our settings using the official implementation of \cite{lee-etal-2022-direct}.\vspace{0.1cm}} 
  \label{t_comp}
  \centering
  \begin{tabular}{l|c|c}
    \toprule
    Method  & En $\rightarrow$ Es & Es $\rightarrow$ En \\
    \midrule
    Without parallel data & & \\
    \quad Translatotron 3$^\clubsuit$ & 13.45 & 14.25 \\
    \quad \textbf{TranSentence (Ours)} & \textbf{18.72} & \textbf{18.24} \\
    \midrule
    With parallel data & & \\
    \quad Translatotron$^\Diamond$ & - & 14.1 \\
    \quad Translatotron 2$^\Diamond$ & - & 30.1 \\
    \quad TranSpeech$^\spadesuit$ & 14.94 & - \\
    \quad S2UT (Transformer encoder)$^\dagger$  & 26.53 & 19.38 \\
    \quad S2UT (Conformer encoder)$^\triangle$ & - & 29.0 \\
    \quad Unity$^\triangle$ & - & 32.3 \\
    \bottomrule
  \end{tabular}
\end{table}

\begin{table}[!t]
  \caption{BLEU scores on the CVSS-C dataset for multilingual speech-to-speech translation by a single model.\vspace{0.1cm}}
  \label{t_multi}
  \centering
  \begin{tabular}{l|cc}
    \toprule
    X  & En $\rightarrow$ X & X $\rightarrow$ En \\
    \midrule
    Es &  18.90 & 18.33  \\
    Fr & 14.69   & 16.59 \\
    \bottomrule
  \end{tabular}
\end{table}

\section{Results}
\subsection{Speech-to-speech translation without parallel speech data}
We compared our model with Translatotron 3 \cite{nachmani2023translatotron}, which is trained with mono-lingual data. For a fair comparison, we conducted experiments with the same dataset. However, while Translatotron 3 \cite{nachmani2023translatotron} used a synthesized dataset based on Common Voice 11 to ensure consistency, we trained with the original dataset. Table \ref{t_comp} exhibited that TranSentence notably surpassed the comparison model. 
\vspace{0.1cm}

\subsection{Multilingual speech-to-speech translation}
We expanded our model to multilingual speech-to-speech translation system. 
In the training phase, we constructed the dataset using non-parallel speech data across three languages: English, French, and Spanish. We then trained a single model to reconstruct the input speech to its original language with additional language ID. In the inference phase, we conditioned on the desired target language ID to generate speech from the speech embedding. 
Table \ref{t_multi} demonstrated that TranSentence with language conditioning can execute multilingual speech-to-speech translation without any performance degradation, even when compared to model that is trained with a single target language dataset.
\vspace{1mm}

\begin{figure}[!t]
  \centering
\centerline{\includegraphics[width=0.90\columnwidth]{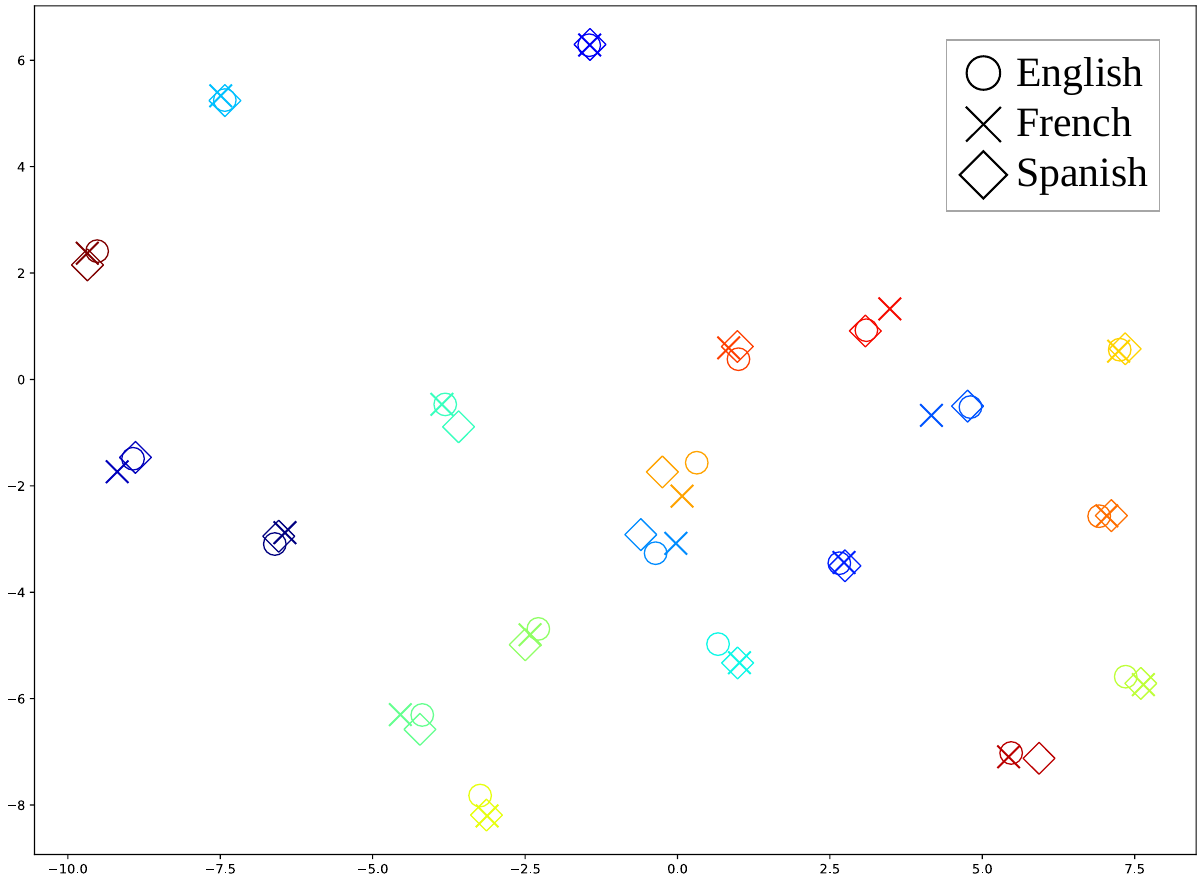}}
\caption{The visualization of language-agnostic sentence-level speech embeddings using t-SNE. Each symbol represents the language of the speech, and speeches with same meanings share the identical color.}
\label{f_tsne} \vspace{-1mm}
\end{figure}

\subsection{Similarity between speech embeddings}
Our proposed method is based on the hypothesis that embeddings, encoded by the language-agnostic sentence-level speech encoder for speech with the same meaning, are similar to each other irrespective of language.
To verify this, we calculated the cosine similarity between speech embeddings from source and target language speech.
For similarity evaluation, we mined language-parallel data for three languages from the CVSS-C dataset, using English translation transcripts as the reference.
As shown in Table \ref{t_sim}, we can confirmed that embeddings encoded from speech with the same meaning are similar to each other, regardless of language. 
Meanwhile, there is room for improvement of speech-to-speech translation by enhancing the language-agnostic sentence-level speech encoder because the embeddings do not yet exactly align with each other. Additionally, when visualizing \cite{Nam_Gur_Choi_Wolf_Lee_2020} the speech embeddings using t-SNE, as shown in Fig. \ref{f_tsne}, we can observe that embeddings with the same meaning cluster together, regardless of language.

We propose an evaluation method for translated speech based on the language-agnostic sentence-level speech encoder. The similarity between speech embeddings of the translated speech and reference speech allows us to evaluate how semantically close the speeches are to each other. The results of the comparison based on cosine similarity are shown in Table \ref{t_eval_sim}.
\vspace{3mm}

\subsection{Ablation study}
We conducted an ablation study to demonstrate the effectiveness of our proposed methods. First, we conducted experiments using different values of $N_{sub}$ to determine the optimal number of sub-embeddings. As indicated in Table \ref{t_able}, speech-to-speech translation performance varies depending on $N_{sub}$. If the $N_{sub}$ is set low, an excessive amount of information is concentrated into a single sub-embedding. Conversely, setting the $N_{sub}$ high leads to excessive feature expansion, resulting in the inclusion of unnecessary information in the sub-embeddings. We also confirmed that refining the sub-embeddings through a semantic encoder enhances the overall performance.
\vspace{0.1cm}

\section{Conclusion}
We proposed TranSentence, a speech-to-speech translation system without language-parallel speech data. 
We successfully introduced language-agnostic sentence-level speech encoding to the S2ST system with representation modeling methods. 
Furthermore, we expanded TranSentence to the multilingual S2ST system, enabling S2ST solely with the speech data of each language without any parallel data. 
We also proposed the evaluation metric for semantic similarity of translated and reference speech through sentence-level speech encoding.
The experimental results exhibited that TranSentence outperformed the comparison model in BLEU scores. 
The limitations of our method include the need for speech-text pair data during the pre-training phase of the speech encoder and the dependency of the translation's performance on the pre-trained speech encoder. Therefore, for future work, we will investigate language-agnostic sentence-level speech encoders that are optimized for generative tasks, rather than for the current mining tasks. 
\vspace{2mm}

\begin{table}[!t]
  \caption{Analysis of sentence embeddings encoded from each language speech via language-agnostic sentence-level speech encoding.\vspace{0.1cm}}
  \label{t_sim}
  \centering
  \begin{tabular}{l|ccc}
    \toprule
    Method & En-Es & En-Fr & Es-Fr \\
    \midrule
    Cosine similarity & 0.9519 & 0.9478 & 0.9344 \\
    \bottomrule
  \end{tabular}
\end{table}

\begin{table}[!t]
  \caption{Cosine Similarity between language-agnostic sentence-level speech embeddings of the translated speech and reference speech. S2UT$^\dagger$ is trained with language-parallel data.\vspace{0.1cm}}
  \label{t_eval_sim}
  \centering
  \begin{tabular}{l|ccc}
    \toprule
    Method & En $\rightarrow$ Es & Es $\rightarrow$ En\\
    \midrule
    Unit vocoded & 0.9564 & 0.9807 \\    
    \midrule
    S2UT$^\dagger$ & \textbf{0.8993} & 0.8513 \\
    TranSentence (Ours) & 0.8732 & \textbf{0.8536} \\
    \bottomrule
  \end{tabular}
\end{table}

\begin{table}[!t]
  \caption{Comparison of BLEU scores for the ablation study on CVSS-C dataset. $N_{sub}$ denotes the number of sub-embeddings.\vspace{0.1cm}} 
  \label{t_able}
  \centering
  \begin{tabular}{l|c|c}
    \toprule
    Method  & $N_{sub}$ & BLEU $(\uparrow)$ \\
    \midrule
    TranSentence (Ours) & 32 & 16.22 \\
     & 16 & \textbf{18.24} \\
     & 8 & 17.17 \\
    \midrule
    w/o semantic encoder & 16 & 16.17 \\
    w/o sub-embeddings & 1 & 14.75 \\
    \bottomrule
  \end{tabular}
\end{table}

\section{Acknowledgements}
This work was partly supported by Institute of Information \& Communications Technology Planning \& Evaluation (IITP) grant funded by the Korea government (MSIT) (No. 2019-0-00079, Artificial Intelligence Graduate School Program (Korea University) and No. 2021-0-02068, Artificial Intelligence Innovation Hub).

\vfill\pagebreak

\ninept
\bibliographystyle{IEEEbib}
\bibliography{refs}

\ninept
\end{document}